\documentclass[letterpaper]{article} 
\usepackage{aaai2027}
\usepackage[hyphens]{url} 
\usepackage{graphicx} 
\usepackage{natbib} 
\usepackage{caption} 
\usepackage{booktabs}
\usepackage{multirow}
\usepackage{colortbl}
\usepackage{amsmath}
\usepackage{amssymb}
\usepackage{algorithm}
\usepackage{algorithmic}
\nocopyright

\newcommand{\method}{\textsc{AlphaG-OPD}}
\newcommand{\pf}{P_F}
\newcommand{\pb}{P_B}
\newcommand{\tb}{\mathcal{L}_{\mathrm{TB}}}
\newcommand{\etb}{\mathcal{L}_{\mathrm{ETB}}}
\newcommand{\opd}{\mathcal{L}_{\mathrm{OPD}}}
\newcommand{\KL}{D_{\mathrm{KL}}}
\definecolor{oursrow}{RGB}{234,244,255}

\title{AlphaG-OPD: Reliability-Gated Sibling Counterfactuals for On-Policy Distillation in Symbolic Alpha Factor Discovery}
\author{Yaoyu Su}
\affiliations{
syy21@tsinghua.org.cn
}

\begin{document}

\maketitle

\begin{abstract}
Symbolic alpha factor discovery can score a completed expression, but it
provides no direct label for the structural decisions that produced it.
Generative flow networks (GFlowNets) preserve a diverse,
reward-proportional distribution over complete
expressions, yet their trajectory-level objective does not compare unchosen
sibling actions at an intermediate state.  We introduce \method, a structural
on-policy distillation framework that turns terminal factor evaluations into
local action guidance.  Its design separates three decisions.  Component I
determines \emph{where to teach} by exposing grammar-valid siblings at partial
abstract-syntax-tree (AST) states visited by the current forward policy.
Component II determines
\emph{what is reliable enough to teach}: it evaluates three supported siblings
under four shared suffixes and admits a KL-bounded target only when their
matched comparisons exhibit sufficient winner agreement and a positive
empirical lower confidence bound (LCB).  Component III determines \emph{how strongly and for
how long to teach} by consolidating accepted targets through bounded replay,
score-indexed expiry, and forward-gradient balancing, without additional
factor evaluations.  Terminal reward, Trajectory Balance, the backward policy,
grammar, and factor-pool rules remain unchanged.  An equal-physical-score
four-arm ablation tests paired teaching, reliability gating, and consolidation.
Across China's CSI300, CSI500, and CSI1000 and the U.S. S\&P 500, the complete
method delivers strong cross-market performance over multiple random seeds.
\end{abstract}

\section{Introduction}
\label{sec:introduction}

Symbolic alpha discovery searches a combinatorial program space, yet a factor
is scored only after its expression is complete.  Genetic programming and
neural symbolic regression explore this space effectively
\citep{autoalpha,cui2021alphaevolve,landajuela2022unified,biggio2021neural},
while REINFORCE and proximal policy optimization (PPO)-style methods directly
optimize expected terminal
quality~\citep{yu2023generating,zhao2025quantfactor,schulman2017proximal}.
GFlowNets provide a complementary strength: rather than return one maximizer,
they learn a reward-proportional distribution and can preserve multiple useful,
nonredundant factor modes~\citep{bengio2023gflownet,chen2025alphasage}.  These
methods solve search and exploration; they do not by themselves identify which
alternative action at a partial expression caused a better terminal outcome.

This distinction between global quality and local credit is consequential in
AST search.  Two trajectories may share the same partial expression but close
it with different operands, operators, or normalization choices.  Their terminal
score entangles the action under study with all of those later decisions.  A
larger search budget can reduce uncertainty about which complete expressions
are valuable, yet it still does not identify which legal action was preferable
at the shared prefix.  Direct local supervision therefore requires a paired
comparison that changes the action while controlling the reachable completion.

On-policy distillation (OPD) supplies precisely this missing interface.  The
evolving student policy
provides the states, and a teacher supplies a local action target at those same
states.  This reduces the state-distribution mismatch of static distillation
and turns feedback on the student's current behavior into a direct probability
update~\citep{gu2024minillm,agarwal2024gkd,song2026opdsurvey}.  In our setting,
TB remains the global search objective; OPD is helpful as a complementary
local-credit channel that can say which next construction should receive more
mass on an AST the current GFlowNet actually visits.

The difficulty is that an unfinished symbolic expression has no external
teacher.  Reusing its eventual terminal score merely repeats TB, independently
completing sibling actions confounds action quality with continuation
difficulty, and repeatedly distilling an unstable preference can be harmful.
Recent OPD studies similarly find that teacher compatibility, local
teachability, and supervision horizon determine whether on-policy guidance
helps or hurts~\citep{li2026rethinkingopd,fu2026revisitingopd,
armandpour2026unmaskingopd,liu2026prefixfade,liang2026adwin}.  Structural OPD
therefore needs matched action comparisons, an explicit reliability rule, and
a bounded optimization budget.

\begin{figure*}[!t]
  \centering
  \includegraphics[width=\textwidth]{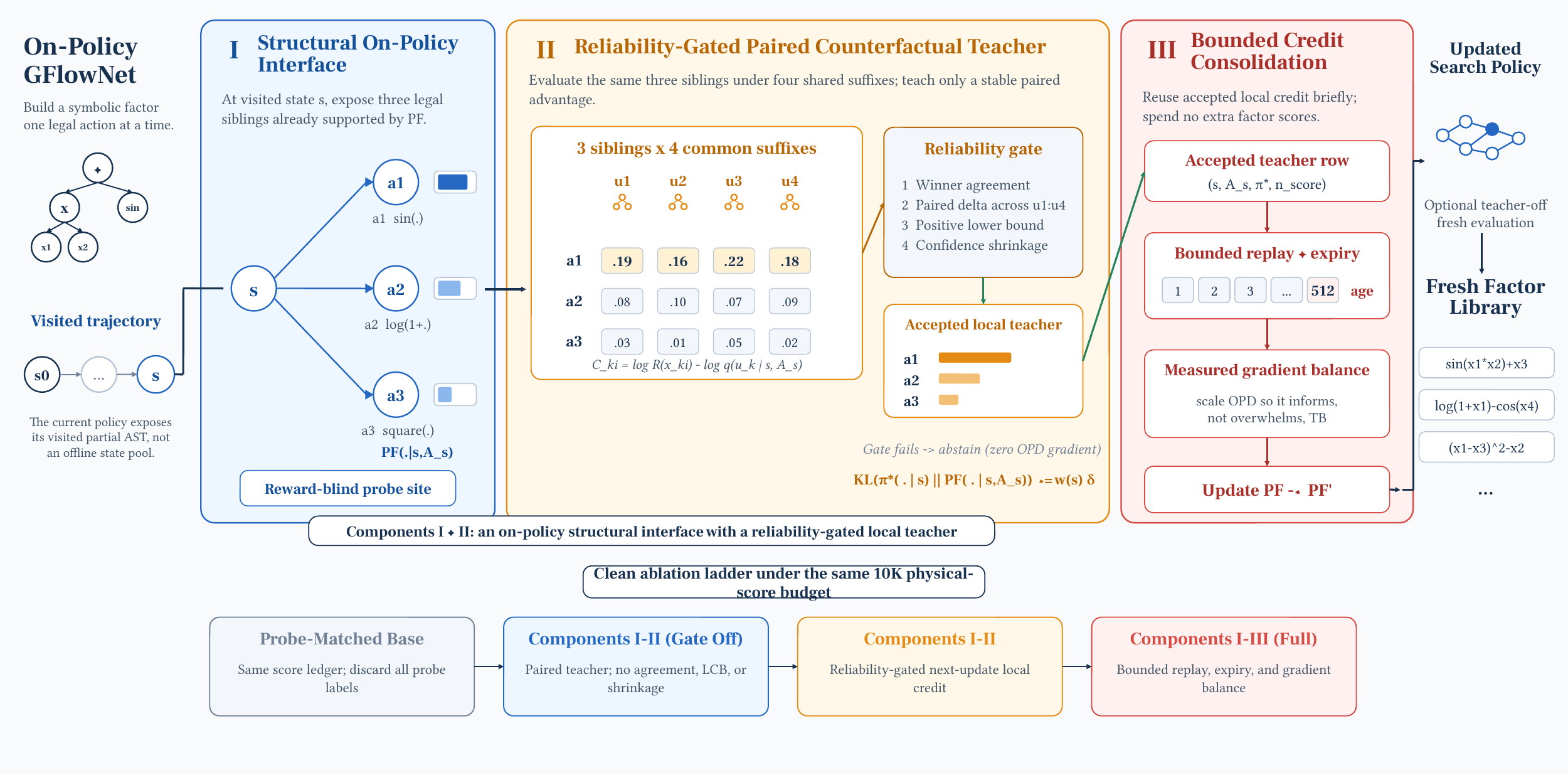}
  \caption{Overview of \method{} through its three questions. Component I
  determines where OPD may teach by exposing supported structural actions at
  an on-policy partial AST. Component II determines what is reliable enough to
  teach by comparing siblings under four common suffixes. Component III
  controls how strongly and for how long accepted credit persists through
  bounded replay and measured forward-gradient balancing. All variants use the
  same physical score budget.}
  \label{fig:method-overview}
\end{figure*}

We introduce \method{}, to our knowledge the first structural OPD framework for
GFlowNet-based symbolic alpha factor discovery.  \method{} converts terminal
factor evaluations into selective local action supervision by comparing
grammar-compatible sibling actions under matched completions at states visited
by the current GFlowNet, then consolidates only accepted credit under a fixed
evaluator budget without replacing Trajectory Balance.  The framework is
organized around three questions: \emph{where may OPD intervene, which local preference is
reliable enough to teach, and how strongly and for how long should accepted
credit influence the policy?}  A reward-blind structural interface, a paired
counterfactual teacher, and bounded consolidation answer these questions in
sequence.  Figure~\ref{fig:method-overview} summarizes the division of labor.

Our contributions are:
\begin{itemize}
    \item \textbf{Where to teach: a structural on-policy interface.}
    We expose grammar-valid sibling decisions at partial AST states visited by
    the evolving forward policy.  State selection is reward-blind, and local
    teaching is restricted to legal actions already supported on the current
    search frontier.
    \item \textbf{What to teach: a reliability-gated paired preference.}
    We evaluate three siblings under four common completions and admit a
    KL-bounded target only when the matched samples agree on the preferred
    action and yield a positive empirical advantage lower bound.  This creates
    a self-generated teacher without an external model or a held-out
    verification-score budget.
    \item \textbf{How accepted credit persists: budget-neutral bounded consolidation.}
    We reuse immutable accepted targets through score-indexed expiry and
    measured forward-gradient balancing.  Replay spends no additional factor
    evaluations, while TB remains the global objective.
\end{itemize}

\section{Related Work}
\label{sec:related}

\subsection{Symbolic Alpha Discovery}

Formulaic alpha mining has progressed from genetic programming and evolutionary
search~\citep{autoalpha,cui2021alphaevolve,AFP} to neural symbolic
regression~\citep{landajuela2022unified,biggio2021neural,shojaee2024llm,
y12,y20,DBLP:conf/iclr/PetersenLMSKK21} and
sequential decision methods.  AlphaGen formulates expression construction as a
Markov decision process (MDP) and searches for synergistic factor
sets~\citep{yu2023generating};
QuantFactor REINFORCE targets lower-variance policy optimization
\citep{zhao2025quantfactor}; and AlphaForge
couples factor generation with dynamic combination~\citep{shi2025alphaforge}.
AlphaAgent regularizes LLM-driven exploration to counteract factor
homogenization and decay~\citep{tang2025alphaagentllmdrivenalphamining} and
serves as a cross-paradigm LLM-agent reference in our comparison.
AlphaSAGE combines a structure-aware encoder with GFlowNet training and is our
primary generative backbone~\citep{chen2025alphasage}.  Relational graph
convolutions, attention, and graph transformers provide complementary
structure-aware encoders~\citep{schlichtkrull2018modeling,vaswani2017attention,
yun2019graph,oono2019graph}, while relation graphs have also been used directly for stock
forecasting~\citep{y23,y24}.  Non-symbolic comparisons include tree ensembles,
latent factor models, and dynamic allocation policies
\citep{chen2016xgboost,ke2017lightgbm,hochreiter1997long,chung2014empirical,
duan2022factorvae,y22,y26}.  These methods
operate over large operator libraries such as Alpha101~\citep{alpha101}, where
many syntactically different trajectories can encode correlated economic
behavior.

\begin{figure}[t]
  \centering
  \includegraphics[width=\columnwidth]{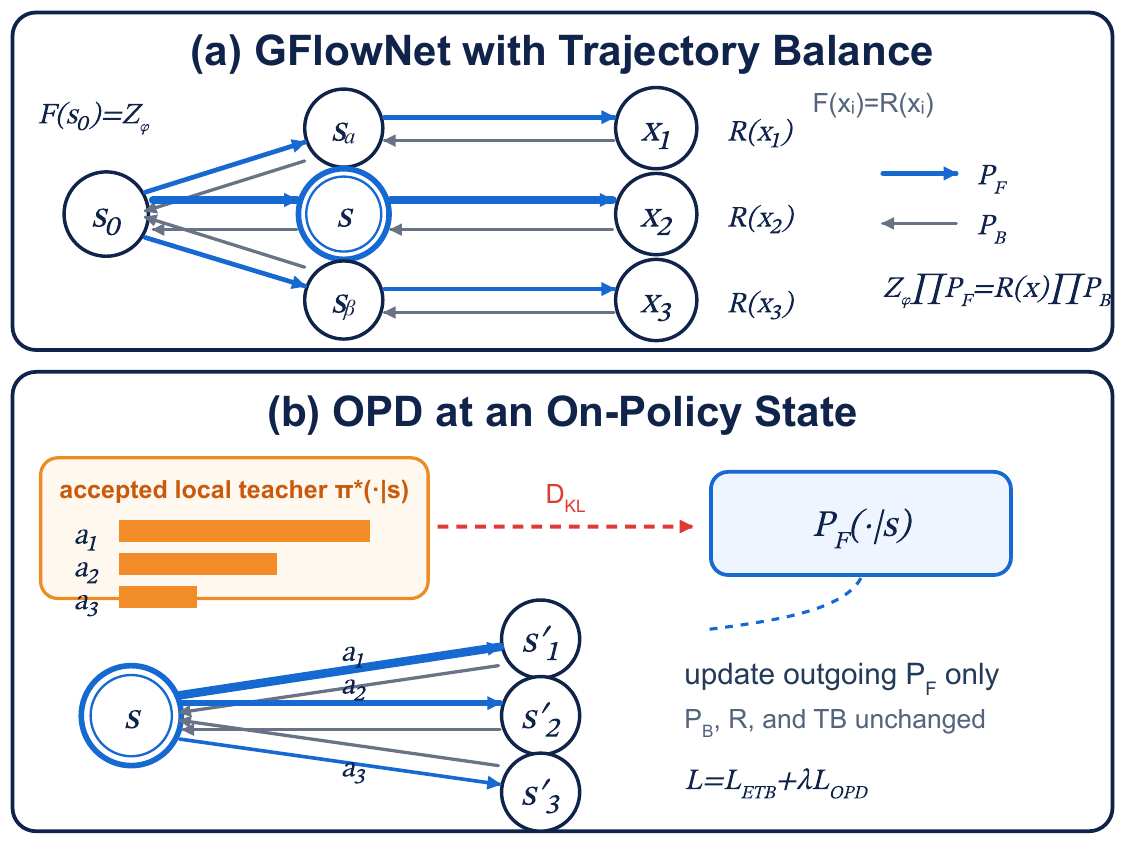}
  \caption{Trajectory-level TB and local OPD. The conceptual state flow $F$
  satisfies $F(s_0)=Z_\phi$ and $F(x)=R(x)$. TB balances $P_F$, $P_B$, and terminal
  reward; at visited $s$, \method{} distills an accepted local sibling teacher
  into outgoing $P_F$ only. $P_B$ and $R$ remain unchanged.}
  \label{fig:tb-opd}
\end{figure}

\subsection{GFlowNets and Local Credit}

GFlowNets sample terminal objects proportionally to a nonnegative reward
\citep{bengio2023gflownet}.  TB enforces a trajectory-level flow identity and
provides a strong terminal-reward objective~\citep{malkin2022trajectory}.
Unlike conventional expected-return policy optimization and constrained-action
implementations~\citep{schulman2017proximal,y6}, this distributional objective supports multiple
high-reward modes and has been useful in diversity-sensitive generation
settings such as biological sequence design~\citep{den}.
Subtrajectory balance supplies additional constraints on partial
trajectories~\citep{madan2023learning}; forward-looking detailed balance uses
partial reward structure to obtain denser updates~\citep{pan2023better}; and
guided or replay-based approaches prioritize informative
trajectories~\citep{shen2023towards}.  Backward-policy optimization can also
improve the credit signal~\citep{gritsaev2024optimizing}.  These objectives
change where a flow constraint is applied.  Our focus is complementary: we
test whether an auxiliary action preference contains information beyond the
terminal objective and whether its direction remains positive across repeated
matched completions.

Figure~\ref{fig:tb-opd} separates the signals: TB remains global, while OPD
only reallocates mass among legal siblings at the visited state.

\subsection{On-Policy Distillation for Structured Search}

Policy distillation transfers a teacher distribution to a student, usually
through KL or cross-entropy~\citep{rusu2015policy}.  MiniLLM derives reverse-KL
on-policy optimization for generative models~\citep{gu2024minillm}, while GKD
queries a teacher on student-generated sequences and unifies on- and off-policy
mixtures with multiple divergences~\citep{agarwal2024gkd}.  Later formulations
extend the interface to context-conditioned teachers~\citep{ye2026opcd} and
organize OPD by feedback source and loss granularity~\citep{song2026opdsurvey}.
On-policy self-distillation instead constructs a privileged-context teacher
from the same model~\citep{zhao2026selfdistilled}.
Objective analyses further separate prefix source from KL direction and revisit
the advantage used by the update~\citep{zhao2026decouplingopd,zhao2026opdplus}.

OPD is not automatically beneficial.  Teacher--student compatibility and
student-prefix reliability can reverse its effect~\citep{li2026rethinkingopd,
fu2026revisitingopd,armandpour2026unmaskingopd}.  Selective variants restrict
teacher influence by rollout, action region, token, or horizon: best-of-$N$ and
step-wise weighting~\citep{zhang2026brts,zhong2026sod}; asymmetric and
trust-region updates~\citep{jia2026aopd,xing2026tropd}; trajectory filtering
and prefix pruning~\citep{li2026fireopd,yang2026pruneopd}; and horizon-aware
supervision~\citep{liu2026prefixfade,liang2026adwin}.  These findings support
selective, state-matched guidance rather than an unconditional auxiliary KL.

They still do not supply a teacher for symbolic search.  We retain the
on-policy interface---the evolving search policy supplies the states---but
replace token logits with structural counterfactuals.  Shared suffixes make
sibling actions comparable; agreement and a paired advantage LCB suppress
unstable targets; and bounded replay determines whether accepted credit can
persist through subsequent TB updates.

\section{Method}
\label{sec:method}

\subsection{Problem Formulation and Design Overview}
\label{sec:setup}

Let $\mathcal{X}$ be the set of valid factor expressions in a typed domain-
specific language.  Expression construction forms a directed acyclic graph
with initial state $s_0$, partial-expression states $s\in\mathcal{S}$, legal
actions $a\in\mathcal{A}(s)$, and deterministic transition
$\mathsf{T}(s,a)$.  A trajectory
$\tau=(s_0,a_1,s_1,\ldots,a_L,s_L=x)$ terminates at expression
$x\in\mathcal{X}$.  The forward policy $\pf^\theta(a\mid s)$ generates the
trajectory; the backward policy $\pb(s\mid s')$ defines reverse flow.

For nonnegative terminal reward $R(x)$, forward-policy parameters $\theta$,
and a separately learned partition function $Z_\phi$, TB minimizes
\begin{equation}
\begin{aligned}
\tb(\tau)=\Bigg[&\log Z_\phi
+\sum_{t=1}^{L}\log \pf^\theta(a_t\mid s_{t-1})-\log R(x)\\
&-\sum_{t=1}^{L}\log \pb(s_{t-1}\mid s_t)\Bigg]^2.
\end{aligned}
\label{eq:tb}
\end{equation}
The evaluator computes $r_{\mathrm{IC}}(x)$, the train-only mean
cross-sectional Pearson information coefficient (IC).  Its nonnegative reward
is $R(x)=\max\{|r_{\mathrm{IC}}(x)|,\exp(-10)\}$.  The implemented backbone is
Entropy-TB,
$\etb=\mathbb{E}_{\tau}[\tb(\tau)]-
\eta_H\mathbb{E}_{\tau}[\sum_{t=0}^{L-1} H(\pf^\theta(\cdot\mid s_t))]$, where
$H$ is categorical entropy, $\eta_H=0.01$, and the entropy temperature is one.
Pool admission, eviction,
backward-policy semantics, this reward, and the Entropy-TB objective remain
unchanged by \method.

\paragraph{Three-stage design.}
At a visited partial state $s$, consider a supported sibling set
$\mathcal{A}_s=\{a_1,\ldots,a_m\}\subseteq\mathcal{A}(s)$.  A local teacher
must estimate how changing only $a$ changes reachable terminal quality while
controlling the completion that follows.  \method{} answers three questions in
sequence.  \emph{Where may OPD intervene?} Component I identifies a legal,
on-policy decision unit.  \emph{What is reliable enough to teach?} Component
II constructs a paired local preference and abstains when its empirical
evidence is weak.  \emph{How much and how long should it teach?} Component III
consolidates only accepted credit under explicit lifetime and gradient bounds.
This decomposition separates eligibility, evidence, and persistence.
Components I--II form the one-use teacher; the full method activates all three
components.  Only $\pf$ receives the auxiliary gradient; terminal reward,
$\pb$, TB, and factor-pool rules are unchanged.

\subsection{Why an On-Policy Distillation Channel}
\label{sec:why-opd}

Existing search objectives remain useful, but their supervision has the wrong
granularity for the local question considered here.  In terminal-reward
symbolic search, expected-return methods
apply a terminal advantage to the actions sampled along one trajectory, with
PPO changing how that trajectory-level signal is clipped and reused.  TB uses
a different, flow-matching objective, but it likewise observes the sampled
trajectory and its terminal reward.  A high-scoring expression therefore
reinforces the actions that occurred; it does not reveal whether an unchosen
legal sibling would have been better at the same partial expression.  Applying
balance constraints to shorter subtrajectories can densify optimization, but
without an additional comparison it does not create that missing action label.

OPD provides exactly this additional interface.  It queries a teacher at states
visited by the evolving student and updates a distribution over several legal
actions, rather than assigning one terminal scalar to the chosen path.  A single
KL update can consequently increase one sibling and decrease another while the
global TB objective continues to preserve reward-proportional, multimodal
search.  The on-policy state source is important: a fixed offline teacher over
states that the current miner rarely visits would reintroduce a state-distribution
mismatch and spend supervision away from the active search frontier.

The interface alone is not sufficient because symbolic search has no external
teacher.  Component II builds a local target from matched terminal
counterfactuals and abstains when the paired evidence is unreliable; Component
III gives accepted sparse targets enough, but not unlimited, optimization
strength.  The division of labor is therefore explicit: TB determines which
complete expressions deserve global probability mass, while OPD redistributes
local forward probability only among compared siblings at a state the current
policy actually encounters.

\subsection{Component I: On-Policy Structural Interface}
\label{sec:interface}

\paragraph{Question 1: Where may OPD intervene?}
The current forward policy samples a syntactically valid trajectory. Candidate
states are partial ASTs from that trajectory, selected without consulting the
probe's terminal scores.  At state $s$, \method{} retains the sampled action
and two alternatives whose current probabilities exceed $10^{-4}$ and whose
Reverse Polish Notation (RPN) transition signatures match.  In particular, operator siblings share
family and arity, while feature, constant, and lag actions are compared only
within their own classes.  This preserves unresolved stack obligations so one
completion grammar can remain legal after every sibling.  A probe is formed
only when such a triple exists; otherwise the method continues ordinary
Entropy-TB without spending a counterfactual score.  The resulting interface is
\begin{equation}
\begin{aligned}
\mathcal{I}(s)&=\left(s,\mathcal{A}_s,
\{p_i\}_{i=1}^{3}\right),\\
\mathcal{A}_s&=\{a_1,a_2,a_3\},\qquad
p_i=\frac{\pf^\theta(a_i\mid s)}
{\sum_{a_j\in\mathcal{A}_s}\pf^\theta(a_j\mid s)}.
\end{aligned}
\label{eq:structural-interface}
\end{equation}
\paragraph{Answer and invariant.}
Component I restricts teaching to states the student actually visits and
actions it can legally take.  The full partial state matters because the same
token may satisfy different typed AST obligations in different contexts.
Component I does not inspect counterfactual scores or factor-pool contents,
change admission rules, or decide which sibling is better.  It determines only
where a local comparison is well defined.

\paragraph{Design trade-off.}
On-policy, type-matched probes reduce coverage but ensure that each comparison
is reachable and grammar-compatible.  Broader probing produces more rows at
the risk of mixing unreachable states or incomparable actions.  Component I
chooses fewer, well-defined teaching contexts.

\subsection{Component II: Reliability-Gated Paired Counterfactual Teacher}
\label{sec:teacher}

\paragraph{Question 2: What is reliable enough to teach?}
The structural interface supplies candidate actions but no preferred action.
Component II creates that preference from matched train-only evaluations and
uses an empirical reliability rule to decide whether any OPD update is
warranted.

\paragraph{Matched completion matrix.}
For each interface $\mathcal{I}(s)$, a reward-blind shared proposal samples
$K=4$ unique suffixes $U=\{u_1,\ldots,u_K\}$ that validly close every sibling.
At each suffix step, the proposal is the normalized geometric mean of the three
current branch policies on their common legal support.  A fixed logit bias
$b_{\mathrm{exit}}=1.5$ is added when exit is common-legal.  Let
$q(u_k\mid s,\mathcal{A}_s)$ be the product of these stepwise proposal
probabilities and $x_{ki}=x(s,a_i,u_k)$ the resulting terminal expression.  The
implemented credit is
\begin{equation}
\mathbf{C}_{ki}=\log R(x_{ki})-
\log q(u_k\mid s,\mathcal{A}_s).
\label{eq:paired-credit}
\end{equation}
This is the log importance contribution of one sampled suffix to the terminal
flow estimator; we use it as a relative sibling credit, not as an unbiased
estimate of log flow.  The
proposal term is common to all siblings in row $k$, so it cancels from both the
row winner and any probability redistribution whose coefficients sum to zero.
A suffix invalid for any sibling is discarded for the entire row, keeping the
$K\times3$ matrix rectangular.  The paired design therefore holds prefix,
continuation, proposal draw, and reachable grammar fixed while changing only
the structural action.

Let $p_i$ be the sibling-normalized probability in
Eq.~\eqref{eq:structural-interface},
$\bar C_i=K^{-1}\sum_k \mathbf{C}_{ki}$, and
$c_i=\bar C_i-\frac{1}{3}\sum_j\bar C_j$.  A provisional teacher is anchored
to the current policy:
\begin{equation}
\widetilde\pi_i(\alpha)=
\frac{p_i\exp(\alpha c_i/\tau_{\mathrm{teach}})}
{\sum_j p_j\exp(\alpha c_j/\tau_{\mathrm{teach}})},
\qquad \KL(\widetilde\pi\|p)\leq\delta,
\label{eq:teacher}
\end{equation}
where a one-dimensional search chooses the nonnegative $\alpha$ whose
Kullback--Leibler (KL) divergence matches the target radius up to numerical
tolerance, $\tau_{\mathrm{teach}}=1$, and $\delta=0.03$.  Here $\KL$ denotes KL
divergence.  Anchoring preserves current-policy support instead of replacing
it with a standalone softmax over noisy action credits.

\paragraph{Paired reliability gate.}
The teacher is used only when the four matched comparisons agree.  Winner
agreement and the per-suffix paired improvement are
\begin{align}
\gamma(s)&=\frac{1}{K}\max_i\sum_{k=1}^{K}
\mathbf{1}\!\left[i=\arg\max_j \mathbf{C}_{kj}\right],
\label{eq:agreement}\\
\Delta_k(s)&=\sum_i\left(\widetilde\pi_i-p_i\right)\mathbf{C}_{ki}.
\label{eq:paired-advantage}
\end{align}
\begin{equation}
\begin{aligned}
\bar\Delta(s)&=\frac{1}{K}\sum_{k=1}^{K}\Delta_k(s),\\
\operatorname{SE}(\Delta)&=
\sqrt{\frac{\sum_{k=1}^{K}(\Delta_k-\bar\Delta)^2}{K(K-1)}},\\
\operatorname{LCB}(s)&=\bar\Delta(s)-z\operatorname{SE}(\Delta).
\end{aligned}
\label{eq:lcb}
\end{equation}
Here $\mathbf{1}[\cdot]$ is the indicator function.
The main method fixes $K=4$, $z=1$, and $\gamma_{\min}=0.75$.  This LCB is an
empirical abstention rule over four paired construction samples, not a 95\%
confidence guarantee.  Component II abstains when credit is constant,
$\gamma<\gamma_{\min}$, or the LCB is nonpositive.  Otherwise it computes
\begin{equation}
w(s)=\min\!\left(1,
\frac{\gamma(s)\operatorname{LCB}(s)}
{\max\{|\bar\Delta|+\operatorname{SE}(\Delta),\epsilon\}}\right),
\qquad \epsilon=10^{-12},
\label{eq:confidence}
\end{equation}
and rebuilds Eq.~\eqref{eq:teacher} with target KL $w(s)\delta$ to obtain
$\pi^*(\cdot\mid s)$.  The implementation recomputes $\Delta_k$ and the LCB for
this shrunken target; reported accepted-row LCBs refer to that final target.
The target is strongest when the preferred action is stable and its paired
improvement is distinguishable from completion noise.

\paragraph{Answer and invariant.}
Component II calls a preference teachable only when the matched rows identify a
stable winner and the candidate redistribution has a positive empirical LCB.
The same four construction suffixes form the teacher and its reliability gate;
the main method uses no held-out verification suffixes.  Each scheduled probe
therefore consumes exactly $3\times4=12$ physical scores, whether the gate
accepts or abstains.  The clean ``Components I--II (Gate Off)'' ablation keeps
the state, siblings, suffixes, temperature, and KL radius fixed while removing
winner agreement, positive-LCB abstention, and confidence shrinkage.  It asks
whether reliability gating adds value beyond paired evaluation alone.

\paragraph{Design trade-off.}
Independent completions cover more continuations but confound action quality
with suffix difficulty.  Four shared suffixes spend coverage to obtain paired
contrasts and an uncertainty estimate.  Abstention then exchanges update
frequency for a smaller set of directionally reliable local targets.

For Algorithm~\ref{alg:alphag-opd}, let $n_{\mathrm{score}}$ denote the
cumulative physical terminal-score count and let $\rho$ denote the target
ratio between the auxiliary and Entropy-TB forward-policy gradient norms.
Component III fixes $\rho=0.10$; Components I--II use a one-update target and
do not apply gradient balancing.

\begin{algorithm}[H]
\footnotesize
\caption{Reliability-Gated Structural OPD}
\label{alg:alphag-opd}
\begin{algorithmic}[1]
\REQUIRE $\pf^\theta$, $\pb$, reward $R$, score ledger, target KL $\delta$,
agreement threshold $\gamma_{\min}$, LCB multiplier $z$, gradient ratio $\rho$
\FOR{each TB episode until the physical score budget is exhausted}
    \STATE Sample a current-policy trajectory; evaluate $R$ and accumulate
    Entropy-TB
    \IF{a sibling probe is scheduled and its 12 scores fit the remaining budget}
        \STATE Select reward-blind visited state $s$ and sibling set
        $\mathcal{A}_s$ from the sampled trajectory
        \STATE Sample four common valid suffixes from $q$
        \STATE Evaluate $\mathbf{C}\in\mathbb{R}^{4\times3}$; construct the
        anchored candidate in Eq.~\eqref{eq:teacher}
        \IF{Eqs.~\eqref{eq:agreement} and~\eqref{eq:lcb} pass}
            \STATE Rebuild $\pi^*$ at radius $w(s)\delta$
            \IF{Component III is enabled}
                \STATE Enqueue $(s,\mathcal{A}_s,\pi^*,n_{\mathrm{score}})$
                for bounded replay
            \ELSE
                \STATE Place the accepted row in the next-update one-use buffer
            \ENDIF
        \ELSE
            \STATE Abstain
        \ENDIF
    \ENDIF
    \STATE At an optimizer step, select pending one-use rows (I--II) or
    unexpired replay rows (I--III)
    \STATE Add Eq.~\eqref{eq:opd-loss}; with Component III, scale it by
    Eq.~\eqref{eq:gradient-balance}; update parameters
\ENDFOR
\ENSURE Score-indexed checkpoints; unchanged $R$, Entropy-TB, $\pb$, and pool
rules
\end{algorithmic}
\end{algorithm}

\renewcommand{\dbltopfraction}{0.58}
\setcounter{dbltopnumber}{1}
\providecommand{\meanstd}[2]{\shortstack{$#1$\\[-2.4pt]{\tiny$(#2)$}}}
\providecommand{\methodfamily}[1]{\textit{#1}}

\begin{table*}[!t]
\centering
\caption{Main comparison by method family. Literature-reference rows are from
AlphaSAGE~\citep{chen2025alphasage}; GFlowNet rows report mean (SD) over three
random seeds. Bold marks the better completed result within each
matched GFlowNet pair.}
\label{tab:main-results}
{\scriptsize
\setlength{\tabcolsep}{1.10pt}
\setlength{\aboverulesep}{0.25ex}
\setlength{\belowrulesep}{0.25ex}
\setlength{\cmidrulesep}{0.20ex}
\renewcommand{\arraystretch}{0.82}
\resizebox{\textwidth}{!}{%
\begin{tabular}{@{}l!{\hspace{2pt}\color{black!35}\vrule width .35pt\hspace{2pt}}l!{\hspace{2pt}\color{black!20}\vrule width .25pt\hspace{2pt}}l*{7}{c}!{\hspace{5pt}\color{black!45}\vrule width .45pt\hspace{5pt}}l!{\hspace{2pt}\color{black!35}\vrule width .35pt\hspace{2pt}}l!{\hspace{2pt}\color{black!20}\vrule width .25pt\hspace{2pt}}l*{7}{c}@{}}
\toprule
Dataset & Family & Method & IC (\%)$\uparrow$ & ICIR$\uparrow$ & RankIC (\%)$\uparrow$ & RankICIR$\uparrow$ & AR (\%)$\uparrow$ & MDD (\%)$\downarrow$ & SR$\uparrow$ &
Dataset & Family & Method & IC (\%)$\uparrow$ & ICIR$\uparrow$ & RankIC (\%)$\uparrow$ & RankICIR$\uparrow$ & AR (\%)$\uparrow$ & MDD (\%)$\downarrow$ & SR$\uparrow$ \\
\midrule
\multirow{10}{*}{CSI300}
& \methodfamily{LLM Agent} & AlphaAgent & 5.10 & 0.325 & 5.60 & 0.329 & 2.16 & 26.9 & 0.65
& \multirow{10}{*}{CSI1000} & \methodfamily{LLM Agent} & AlphaAgent & 7.20 & 0.579 & 8.90 & 0.712 & 5.51 & 20.5 & 1.01 \\
\cmidrule(lr){2-10}\cmidrule(lr){12-20}
& \multirow{4}{*}{\methodfamily{Conventional}} & MLP        & 2.00 & 0.158 & 1.90 & 0.142 & 3.54 & 20.9 & 0.68
& & \multirow{4}{*}{\methodfamily{Conventional}} & MLP        & 4.80 & 0.384 & 6.90 & 0.621 & 3.22 & 25.7 & 0.47 \\
& & LightGBM   & 1.10 & 0.124 & 0.60 & 0.064 & 2.61 & 18.5 & 0.53
& & & LightGBM   & 6.70 & 0.501 & 8.30 & 0.656 & 4.98 & 22.7 & 0.98 \\
& & XGBoost    & 3.10 & 0.243 & 3.30 & 0.248 & 5.40 & 17.5 & 1.26
& & & XGBoost    & 6.20 & 0.498 & 8.60 & 0.695 & 4.72 & 23.5 & 0.91 \\
& & GP         & 2.60 & 0.215 & 2.80 & 0.216 & 6.80 & 17.6 & 1.55
& & & GP         & 5.80 & 0.474 & 7.90 & 0.657 & 4.32 & 24.7 & 0.67 \\
\cmidrule(lr){2-10}\cmidrule(lr){12-20}
& \multirow{3}{*}{\methodfamily{Neural/RL}} & AlphaGen   & 5.80 & 0.414 & 5.70 & 0.360 & 4.00 & 22.6 & 0.76
& & \multirow{3}{*}{\methodfamily{Neural/RL}} & AlphaGen   & 7.10 & 0.540 & 9.20 & 0.713 & 5.27 & 24.0 & 0.92 \\
& & AlphaQCM   & 4.30 & 0.262 & 4.20 & 0.246 & 1.95 & 24.8 & 0.36
& & & AlphaQCM   & 6.50 & 0.453 & 10.70 & 0.682 & 7.12 & 20.6 & 1.31 \\
& & AlphaForge & 4.10 & 0.259 & 5.20 & 0.306 & 3.90 & 21.9 & 0.88
& & & AlphaForge & 7.10 & 0.537 & 9.50 & 0.742 & 6.07 & 21.1 & 1.06 \\
\cmidrule(lr){2-10}\cmidrule(lr){12-20}
& \multirow{2}{*}{\methodfamily{GFlowNet}} & AlphaSAGE
& \meanstd{5.81}{1.03}
& \meanstd{0.4177}{0.0686}
& \meanstd{7.50}{1.69}
& \meanstd{0.5195}{0.0803}
& \meanstd{6.146}{3.787}
& \meanstd{15.140}{1.397}
& \meanstd{1.413}{0.987}
& & \multirow{2}{*}{\methodfamily{GFlowNet}} & AlphaSAGE
& \meanstd{\mathbf{6.04}}{0.28}
& \meanstd{0.4781}{0.0268}
& \meanstd{\mathbf{8.44}}{1.15}
& \meanstd{0.6088}{0.0404}
& \meanstd{\mathbf{4.011}}{0.256}
& \meanstd{\mathbf{24.804}}{3.883}
& \meanstd{\mathbf{0.669}}{0.036} \\
& & \cellcolor{oursrow}\textbf{\method{} (ours)}
& \cellcolor{oursrow}\meanstd{\mathbf{6.73}}{0.74}
& \cellcolor{oursrow}\meanstd{\mathbf{0.4479}}{0.0270}
& \cellcolor{oursrow}\meanstd{\mathbf{8.78}}{1.13}
& \cellcolor{oursrow}\meanstd{\mathbf{0.5408}}{0.0283}
& \cellcolor{oursrow}\meanstd{\mathbf{8.382}}{2.222}
& \cellcolor{oursrow}\meanstd{\mathbf{14.184}}{1.562}
& \cellcolor{oursrow}\meanstd{\mathbf{2.080}}{0.615}
& & & \cellcolor{oursrow}\textbf{\method{} (ours)}
& \cellcolor{oursrow}\meanstd{5.89}{0.33}
& \cellcolor{oursrow}\meanstd{\mathbf{0.4894}}{0.0347}
& \cellcolor{oursrow}\meanstd{7.74}{0.86}
& \cellcolor{oursrow}\meanstd{\mathbf{0.6298}}{0.0715}
& \cellcolor{oursrow}\meanstd{3.155}{0.363}
& \cellcolor{oursrow}\meanstd{25.271}{2.513}
& \cellcolor{oursrow}\meanstd{0.534}{0.081} \\
\midrule
\multirow{10}{*}{CSI500}
& \methodfamily{LLM Agent} & AlphaAgent & 5.30 & 0.396 & 6.50 & 0.495 & 1.82 & 22.4 & 0.36
& \multirow{10}{*}{S\&P 500} & \methodfamily{LLM Agent} & AlphaAgent & 4.80 & 0.479 & 3.30 & 0.315 & 18.66 & 5.7 & 6.27 \\
\cmidrule(lr){2-10}\cmidrule(lr){12-20}
& \multirow{4}{*}{\methodfamily{Conventional}} & MLP        & 1.70 & 0.185 & 2.00 & 0.233 & 1.56 & 24.3 & 0.27
& & \multirow{4}{*}{\methodfamily{Conventional}} & MLP        & 3.50 & 0.287 & 2.00 & 0.143 & 12.85 & 5.6 & 3.35 \\
& & LightGBM   & 2.40 & 0.305 & 2.10 & 0.264 & 4.61 & 17.5 & 0.89
& & & LightGBM   & 2.30 & 0.196 & 1.80 & 0.165 & 11.11 & 5.1 & 4.22 \\
& & XGBoost    & 3.90 & 0.365 & 5.20 & 0.528 & 5.50 & 17.1 & 1.15
& & & XGBoost    & 1.60 & 0.159 & 2.60 & 0.168 & 13.25 & 8.3 & 3.61 \\
& & GP         & 1.40 & 0.238 & 2.20 & 0.233 & 3.04 & 19.4 & 0.56
& & & GP         & 3.20 & 0.308 & 0.20 & 0.016 & 13.39 & 13.0 & 3.15 \\
\cmidrule(lr){2-10}\cmidrule(lr){12-20}
& \multirow{3}{*}{\methodfamily{Neural/RL}} & AlphaGen   & 3.20 & 0.270 & 3.10 & 0.230 & 1.15 & 32.4 & 0.19
& & \multirow{3}{*}{\methodfamily{Neural/RL}} & AlphaGen   & 4.40 & 0.396 & 1.30 & 0.127 & 10.31 & 5.5 & 3.96 \\
& & AlphaQCM   & 4.80 & 0.378 & 7.30 & 0.546 & 4.06 & 24.0 & 0.75
& & & AlphaQCM   & 3.80 & 0.262 & 1.00 & 0.071 & 13.86 & 13.0 & 3.30 \\
& & AlphaForge & 5.30 & 0.345 & 8.30 & 0.600 & 4.18 & 16.7 & 0.93
& & & AlphaForge & 3.90 & 0.422 & 3.10 & 0.324 & 17.24 & 5.0 & 6.30 \\
\cmidrule(lr){2-10}\cmidrule(lr){12-20}
& \multirow{2}{*}{\methodfamily{GFlowNet}} & AlphaSAGE
& \meanstd{1.97}{1.16}
& \meanstd{0.1721}{0.0950}
& \meanstd{2.72}{1.98}
& \meanstd{0.2352}{0.1691}
& \meanstd{-0.778}{2.020}
& \meanstd{25.040}{3.286}
& \meanstd{-0.139}{0.361}
& & \multirow{2}{*}{\methodfamily{GFlowNet}} & AlphaSAGE
& \meanstd{\mathbf{0.89}}{1.26}
& \meanstd{\mathbf{0.0921}}{0.1314}
& \meanstd{1.19}{0.93}
& \meanstd{\mathbf{0.1103}}{0.0950}
& \meanstd{9.213}{1.809}
& \meanstd{26.136}{0.367}
& \meanstd{1.795}{0.319} \\
& & \cellcolor{oursrow}\textbf{\method{} (ours)}
& \cellcolor{oursrow}\meanstd{\mathbf{4.71}}{1.15}
& \cellcolor{oursrow}\meanstd{\mathbf{0.3369}}{0.0547}
& \cellcolor{oursrow}\meanstd{\mathbf{6.81}}{1.81}
& \cellcolor{oursrow}\meanstd{\mathbf{0.4671}}{0.1114}
& \cellcolor{oursrow}\meanstd{\mathbf{2.968}}{1.308}
& \cellcolor{oursrow}\meanstd{\mathbf{18.535}}{1.531}
& \cellcolor{oursrow}\meanstd{\mathbf{0.577}}{0.237}
& & & \cellcolor{oursrow}\textbf{\method{} (ours)}
& \cellcolor{oursrow}\meanstd{0.56}{0.73}
& \cellcolor{oursrow}\meanstd{0.0519}{0.0677}
& \cellcolor{oursrow}\meanstd{\mathbf{1.20}}{0.74}
& \cellcolor{oursrow}\meanstd{0.1043}{0.0619}
& \cellcolor{oursrow}\meanstd{\mathbf{9.398}}{0.627}
& \cellcolor{oursrow}\meanstd{\mathbf{25.530}}{1.418}
& \cellcolor{oursrow}\meanstd{\mathbf{1.882}}{0.107} \\
\bottomrule
\end{tabular}}
}
\end{table*}

\begin{table*}[t]
\centering
\caption{CSI500 additive ablation under the same 10K physical-score
budget and frozen Validation-Sign Equal-50 evaluator.  Validation Pearson IC
fixes each factor's sign at magnitude $1/50$ before the untouched test.}
\label{tab:full-system}
{\scriptsize
\setlength{\tabcolsep}{3.05pt}
\renewcommand{\arraystretch}{1.08}
\resizebox{\textwidth}{!}{%
\begin{tabular}{@{}lccccrrrrrrr@{}}
\toprule
Method & I: Interface & Paired Teacher & Reliability Gate & III: Consolidation &
IC (\%)$\uparrow$ & ICIR$\uparrow$ & RankIC (\%)$\uparrow$ &
RankICIR$\uparrow$ & AR (\%)$\uparrow$ & SR$\uparrow$ & MDD (\%)$\downarrow$ \\
\midrule
Base Model
& -- & -- & -- & --
& 2.175 & 0.182282 & 3.200 & 0.277667 & -1.038 & -0.209 & 24.571 \\
Components I--II (Gate Off)
& \checkmark & \checkmark & -- & --
& 4.040 & 0.342000 & 4.850 & 0.395800 & 0.920 & 0.164 & 22.860 \\
Components I--II (Gate On)
& \checkmark & \checkmark & \checkmark & --
& 3.730 & 0.327100 & 5.520 & 0.453500 & 1.320 & 0.249 & 21.820 \\
\rowcolor{oursrow}
\textbf{\method{} (I--III)}
& \checkmark & \checkmark & \checkmark & \checkmark
& \textbf{6.018} & \textbf{0.399784} & \textbf{8.822} &
\textbf{0.595058} & \textbf{4.384} & \textbf{0.835} & \textbf{17.417} \\
\bottomrule
\end{tabular}}
}
\end{table*}

\subsection{Component III: Bounded Credit Consolidation}
\label{sec:replay}

\paragraph{Question 3: How much, and for how long?}
Components I--II place each accepted teacher in a one-use buffer consumed at the
next optimizer update; multiple accepted rows can therefore share one PF update.
This pending buffer uses the same 512-row capacity as the replay queue.
Component III instead stores the state, sibling actions, target probabilities,
and creation score count $n_{\mathrm{score}}$ in a queue of at most 512 rows.
Rows expire after
1,000 physical terminal evaluations; replay consumes no new factor score and
never recomputes the stored target from a later policy.

For active rows $\mathcal{B}$, define the current sibling-normalized student as
$\pf^\theta(a\mid s,\mathcal{A}_s)=\pf^\theta(a\mid s)/
\sum_{a'\in\mathcal{A}_s}\pf^\theta(a'\mid s)$.  The auxiliary objective is
forward KL from the teacher to this distribution:
\begin{equation}
\begin{aligned}
\opd(\theta)=\frac{1}{|\mathcal{B}|}
&\sum_{(s,\mathcal{A}_s,\pi^*)\in\mathcal{B}}\\[-2pt]
&\KL\!\left(\pi^*(\cdot\mid s)\,\|\,
\pf^\theta(\cdot\mid s,\mathcal{A}_s)\right).
\end{aligned}
\label{eq:opd-loss}
\end{equation}
The KL is computed after both distributions are renormalized on the stored
sibling set.  It therefore changes the relative probabilities of the compared
actions without assigning labels to unrelated legal actions at the same state.
Forward KL is appropriate here because the accepted local teacher, not a new sample
from the student, defines which supported actions must retain probability mass.
The original TB loss remains responsible for states and actions that have no
active accepted row.

Sparse supervision can otherwise be numerically irrelevant compared with
Entropy-TB.
At each optimizer update, \method{} measures the forward-policy gradient norms
$g_{\mathrm{ETB}}=\|\nabla_\theta\etb\|_2$ and
$g_{\mathrm{OPD}}=\|\nabla_\theta\opd\|_2$, then sets
\begin{equation}
\lambda=\min\left(\lambda_{\max},
\rho\frac{g_{\mathrm{ETB}}}{g_{\mathrm{OPD}}+\epsilon}\right),
\qquad
\mathcal{L}=\etb+\lambda\opd,
\label{eq:gradient-balance}
\end{equation}
where $\rho=0.10$ is the target OPD-to-Entropy-TB gradient ratio on $\pf$,
$\lambda_{\max}=10^4$, and $\epsilon=10^{-12}$.  The auxiliary gradient is
exactly zero when no accepted row is active.  All other auxiliary guides are
disabled in the registered arms, so the reference gradient in
Eq.~\eqref{eq:gradient-balance} is the Entropy-TB forward-policy gradient.

Replay age is measured in terminal score calls rather than optimizer steps.
This choice makes freshness comparable across arms even when probing changes
the number of updates between ordinary trajectories.  The stored target is not
silently recomputed from the later policy: doing so would change the teaching
claim after acceptance.  Expiration instead makes the intervention explicit---
the row teaches the reliability-gated preference for a bounded period and then
returns control entirely to current TB learning.

\paragraph{Answer and invariant.}
Component III controls duration through score-indexed expiry and strength
through measured gradient balancing.  It cannot reverse or recompute the
accepted preference, request another factor evaluation, or label actions
outside the stored sibling set.  When no accepted row is active, its auxiliary
gradient is exactly zero; when replay is active, Eq.~\eqref{eq:gradient-balance}
bounds its influence relative to TB.  Thus TB remains the global search
objective throughout training.

\paragraph{Design trade-off.}
One-use supervision can be too weak relative to TB, whereas unlimited replay
becomes stale and loses its connection to the current policy.  Capping row age
by physical score count preserves freshness, while balancing the gradient that
actually reaches $\pf$ gives each accepted preference a controlled amount of
influence.  The Components I--II versus Components I--III comparison isolates
this consolidation increment because both arms use the same teacher-construction
rule; their realized states and accepted rows may differ as their policies
diverge.

\subsection{Ablation Controls and Physical Score Accounting}
\label{sec:controls}

A sibling probe evaluates every supported action under every common suffix.
For $m$ actions and $K$ shared suffixes, its direct evaluator cost is
\begin{equation}
N_{\mathrm{score}}^{\mathrm{probe}}=mK,
\label{eq:probe-cost}
\end{equation}
which is $3\times4=12$ in Component II.  Every call enters the physical-score
ledger even when the gate abstains; the same ledger indexes training,
checkpoints, and replay expiry.  Here $N_{\mathrm{score}}^{\mathrm{probe}}$ is
the per-probe call count, whereas $n_{\mathrm{score}}$ is the cumulative
ledger position stored with an accepted row.

The registered 10K ablation holds the total physical-score budget fixed.
\textsc{Base Model} spends that budget on ordinary generation.  Each OPD
arm spends $8{,}008$ ordinary and $1{,}992$ probe evaluations.
\textsc{Components I--II (Gate Off)} introduces the structural interface and
$4\times3$ paired score matrix without agreement, LCB abstention, or confidence
shrinkage.  \textsc{Components I--II} keeps the same allocation and restores
those gates, then consumes accepted targets once at the next update;
\textsc{Components I--III} changes only their persistence and strength.  Thus
Base Model versus Gate-Off tests the cost-aware paired-teaching intervention, while
the two later adjacent contrasts isolate reliability gating and bounded
consolidation.

\section{Experiments}
\label{sec:experiments}

We evaluate reliability, factor-pool transmission, and portfolio conversion on
the Chinese A-share CSI300/500/1000 universes and the U.S. S\&P 500.  Following
Qlib~\citep{y25}, the target is the 20-day forward close return
\texttt{Ref(close,-20)/close-1}.  China uses
2010--2020/2021/2022--2024 and the U.S. uses 2010--2016/2017/2018--2020 for
train/validation/test.  Matched arms use multiple random seeds, 10K physical scores,
and otherwise frozen rules.  Table~\ref{tab:main-results} uses Validation-Sign
Equal-50: validation Pearson IC fixes each factor's sign at magnitude $1/50$
before one untouched test; S\&P 500 portfolio metrics are long-only.
Table~\ref{tab:full-system} uses the same frozen Validation-Sign Equal-50
evaluator for the CSI500 additive ablation.

IC/RankIC are Pearson/Spearman correlations in percent; ICIR/RankICIR are
unitless.  AR, MDD, and SR denote annualized return, positive maximum drawdown
magnitude, and Sharpe ratio.
Parentheses report sample standard deviation across multiple random seeds, and
bold marks the better completed value within each matched GFlowNet pair.

\section{Results and Analysis}
\label{sec:results}

\begin{figure}[!t]
  \centering
  \includegraphics[width=0.96\columnwidth]{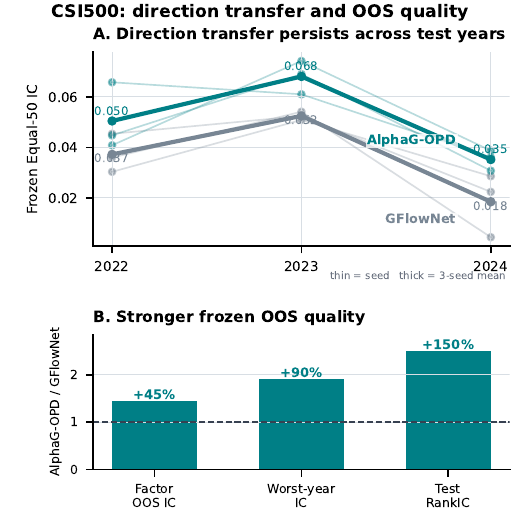}
  \caption{CSI500 direction transfer and frozen OOS quality across multiple random
  seeds.  (A) Directions fixed from 2021 Pearson IC remain positive in every
  2022--2024 test year.  (B) \method{} improves factor OOS IC, worst-year
  Equal-50 IC, and test RankIC by $45/90/150\%$.}
  \label{fig:csi500-quality-redundancy}
\end{figure}

\paragraph{Main comparison.}
Across the completed multi-seed results, \method{} improves all seven CSI300/500
means over AlphaSAGE.  CSI300 IC/RankIC rise from $5.81/7.50\%$ to
$6.73/8.78\%$, AR/Sharpe from $6.146\%/1.413$ to $8.382\%/2.080$, and MDD falls
from $15.140\%$ to $14.184\%$; CSI500 return moves from $-0.778\%$ to $2.968\%$.
Figure~\ref{fig:csi500-quality-redundancy} shows persistent direction transfer
and stronger frozen OOS quality in every test year.  On CSI1000, \method{} improves ICIR and
RankICIR, demonstrating stronger temporal consistency in a broader universe.
On S\&P 500, it improves RankIC, annualized return, Sharpe, and drawdown.
Together, these outcomes connect stronger ranking consistency with stronger
portfolio conversion across distinct market scales.  The multi-seed evidence
demonstrates gains at both factor and portfolio levels.

\paragraph{Component ablation.}
Table~\ref{tab:full-system} shows the staged effect of the three components on
the fixed CSI500 ablation test.  Components I--II with Gate Off already improve
the Base Model.  Enabling the reliability gate shifts the gain toward the
decision-relevant metrics, further improving RankIC, return, Sharpe, and MDD.
Adding Component III then improves all seven metrics over Gate On, yielding
$8.822\%$ RankIC, $4.384\%$ return, $0.835$ Sharpe, and $17.417\%$ MDD.  The
ablation therefore supports the intended progression from paired local credit,
through selective teaching, to bounded consolidation.  This confirms cumulative
component contributions.

\section{Conclusion}
\label{sec:conclusion}

\method{} converts terminal factor evaluations into reliable local structural
guidance while preserving the global TB objective.  Its structural interface,
paired teacher, reliability gate, and bounded consolidation progressively
strengthen decision-relevant factor quality under the same 10K evaluator
budget.  Across four Chinese and U.S. equity markets, the method delivers
comprehensive CSI300/500 gains, stronger CSI1000 information-ratio consistency,
and improved S\&P 500 risk-adjusted performance.  This establishes
budget-efficient structural alpha discovery.  By adding
action-level supervision exactly where the current policy searches, \method{}
turns sparse terminal feedback into reusable structural knowledge.  The same
design operates across multiple universes without changing the grammar,
terminal reward, factor-pool construction, or global flow objective.

\clearpage
\bibliography{references}

\clearpage
\appendix
\setcounter{secnumdepth}{2}
\setcounter{equation}{0}
\setcounter{table}{0}
\setcounter{figure}{0}
\renewcommand{\theequation}{\Alph{section}.\arabic{equation}}
\renewcommand{\thetable}{\Alph{section}.\arabic{table}}
\renewcommand{\thefigure}{\Alph{section}.\arabic{figure}}
\section{Detailed Derivations and Reproduction Contract}
\label{app:details}

This appendix gives the complete notation, mathematical derivations,
physical-score accounting, implementation configuration, data splits, and
evaluation formulas for \method{}.  It expands each of the three components
from the main paper and specifies how every reported factor and portfolio
metric is computed.

\subsection{Notation and Search Process}

Table~\ref{tab:notation} collects the notation used in the main paper and this
supplement.  A state is a typed partial abstract syntax tree (AST), represented
operationally by its Reverse Polish Notation construction state and unresolved
stack obligations.  A deterministic grammar transition maps a legal action to
the next state.  A terminal state is a complete symbolic factor expression.

\begin{table}[t]
\centering
\small
\caption{Core notation.}
\label{tab:notation}
\begin{tabular}{ll}
\toprule
Symbol & Meaning \\
\midrule
$s_0,s,x$ & root, partial state, and terminal expression \\
$\mathcal{A}(s)$ & grammar-valid actions at $s$ \\
$\mathcal{A}_s$ & three supported, type-matched sibling actions \\
$\pf^\theta,\pb$ & forward and backward policies \\
$R(x)$ & nonnegative train-only terminal reward \\
$Z_\phi$ & learned GFlowNet partition function \\
$u_k$ & $k$th shared grammar-valid completion suffix \\
$q(u_k\mid s,\mathcal{A}_s)$ & reward-blind shared suffix proposal \\
$\mathbf C_{ki}$ & credit for suffix $k$ and sibling $i$ \\
$p_i$ & current policy restricted to $\mathcal{A}_s$ \\
$\widetilde\pi,\pi^*$ & provisional and accepted local teachers \\
$\gamma$ & matched-suffix winner agreement \\
$\Delta_k$ & paired improvement on suffix $k$ \\
$n_{\mathrm{score}}$ & cumulative physical terminal-score count \\
\bottomrule
\end{tabular}
\end{table}

A trajectory $\tau=(s_0,a_1,s_1,\ldots,a_L,s_L=x)$ follows
$s_t=\mathsf T(s_{t-1},a_t)$.  For terminal reward $R(x)>0$, Trajectory
Balance (TB) minimizes
\begin{equation}
\begin{aligned}
\tb(\tau)=\Bigg[&\log Z_\phi
+\sum_{t=1}^{L}\log\pf^\theta(a_t\mid s_{t-1})\\
&-\log R(x)
-\sum_{t=1}^{L}\log\pb(s_{t-1}\mid s_t)\Bigg]^2.
\end{aligned}
\label{eq:supp-tb}
\end{equation}
At zero TB residual, exponentiating the expression inside the square gives
\begin{equation}
Z_\phi\prod_{t=1}^{L}\pf^\theta(a_t\mid s_{t-1})
=R(x)\prod_{t=1}^{L}\pb(s_{t-1}\mid s_t),
\end{equation}
which is the trajectory-level flow constraint.  The implemented backbone adds
an entropy bonus,
\begin{equation}
\begin{aligned}
\etb={}&\mathbb E_\tau[\tb(\tau)]\\
&-\eta_H\,\mathbb E_\tau\!\left[\sum_{t=0}^{L-1}
H(\pf^\theta(\cdot\mid s_t))\right],
\quad \eta_H=0.01.
\end{aligned}
\label{eq:supp-etb}
\end{equation}
The terminal evaluator computes the mean cross-sectional Pearson IC on the
training period and uses
$R(x)=\max\{|r_{\mathrm{IC}}(x)|,\exp(-10)\}$.  Equations
\eqref{eq:supp-tb}--\eqref{eq:supp-etb} provide the global flow objective, while
the three components below provide local structural credit to $\pf^\theta$.

\section{Component I: On-Policy Structural Interface}

The intervention point is a state sampled by the current forward policy.  This
on-policy restriction aligns the teacher's state distribution with the states
on which the student is currently making decisions.  At a candidate state,
the sampled action and two alternatives are retained only when all three:
\begin{enumerate}
    \item have current forward probability above $10^{-4}$;
    \item belong to the same action class, operator family, and arity; and
    \item induce the same unresolved construction signature.
\end{enumerate}
These conditions ensure that one suffix can legally complete every sibling.
The interface records
\begin{equation}
\mathcal I(s)=\left(s,\mathcal A_s,\{p_i\}_{i=1}^{3}\right),\qquad
p_i=\frac{\pf^\theta(a_i\mid s)}
{\sum_{a_j\in\mathcal A_s}\pf^\theta(a_j\mid s)}.
\label{eq:supp-interface}
\end{equation}
Renormalization is important: OPD redistributes probability only inside the
stored sibling set.  It neither labels nor directly constrains other legal
actions at $s$.  If no legal supported triple exists, the episode proceeds
with Entropy-TB and no counterfactual evaluator calls are spent.

\section{Component II: Reliability-Gated Paired Teacher}

\subsection{Matched completion proposal}

For a retained interface, the method samples $K=4$ unique suffixes that are
valid after every sibling.  At suffix step $t$, let
$\mathcal V_t$ be the common legal action support and let
$\ell_{it}(b)$ denote the forward logit of branch $i$ for action $b$.  The
reward-blind shared proposal is
\begin{equation}
\begin{aligned}
q_t(b)\propto{}&\mathbf 1[b\in\mathcal V_t]\\
&\times\exp\!\left(\frac{1}{3}\sum_{i=1}^{3}\ell_{it}(b)
+b_{\mathrm{exit}}\mathbf 1[b=\mathrm{exit}]\right),\\
& b_{\mathrm{exit}}=1.5.
\end{aligned}
\label{eq:supp-proposal-step}
\end{equation}
The suffix probability is the product of stepwise probabilities,
$q(u_k\mid s,\mathcal A_s)=\prod_t q_t(u_{kt})$.  The proposal is constructed
without evaluating terminal rewards.  If a sampled suffix becomes invalid for
any branch, the entire matched row is discarded rather than selectively
replacing one branch.

For terminal expression $x_{ki}=x(s,a_i,u_k)$, the relative credit is
\begin{equation}
\mathbf C_{ki}=\log R(x_{ki})-\log q(u_k\mid s,\mathcal A_s).
\label{eq:supp-credit}
\end{equation}
Within row $k$, the proposal term is identical for all siblings.  Hence
\begin{equation}
\mathbf C_{ki}-\mathbf C_{kj}
=\log R(x_{ki})-\log R(x_{kj}),
\label{eq:supp-cancel}
\end{equation}
and the row winner is unaffected by suffix sampling probability.  More
generally, for coefficients $d_i$ satisfying $\sum_i d_i=0$,
\begin{equation}
\sum_i d_i\mathbf C_{ki}=\sum_i d_i\log R(x_{ki}).
\end{equation}
This cancellation makes the comparison depend on sibling reward differences
under the same completion.

\subsection{Anchored exponential tilt}

Average each sibling's matched credit,
$\bar C_i=K^{-1}\sum_k\mathbf C_{ki}$, and center it as
$c_i=\bar C_i-\frac13\sum_j\bar C_j$.  The provisional teacher is
\begin{equation}
\widetilde\pi_i(\alpha)=
\frac{p_i\exp(\alpha c_i/\tau)}
{\sum_j p_j\exp(\alpha c_j/\tau)},
\qquad \tau=1.
\label{eq:supp-teacher}
\end{equation}
This is an exponential tilt of the current sibling distribution, rather than a
new distribution that ignores current support.  At $\alpha=0$,
$\widetilde\pi=p$.  Increasing $\alpha$ moves mass toward larger centered
credit.  The implementation performs a one-dimensional search for the largest
nonnegative tilt satisfying
\begin{equation}
\KL(\widetilde\pi(\alpha)\|p)\leq\delta,
\qquad \delta=0.03.
\label{eq:supp-kl-radius}
\end{equation}
Thus every accepted local target remains in an explicit trust region around
the current policy.

\subsection{Agreement, paired improvement, and abstention}

The suffix-level winner agreement is
\begin{equation}
\gamma(s)=\frac1K\max_i\sum_{k=1}^{K}
\mathbf1\!\left[i=\arg\max_j\mathbf C_{kj}\right].
\label{eq:supp-agreement}
\end{equation}
With $K=4$ and $\gamma_{\min}=0.75$, at least three of the four matched
suffixes must identify the same preferred sibling.  For a candidate teacher,
the paired improvement on row $k$ is
\begin{equation}
\Delta_k=\sum_i(\widetilde\pi_i-p_i)\mathbf C_{ki}.
\label{eq:supp-delta}
\end{equation}
Because $\sum_i(\widetilde\pi_i-p_i)=0$, the shared proposal term in
Eq.~\eqref{eq:supp-credit} cancels from $\Delta_k$.  Positive $\Delta_k$ means
that the proposed redistribution assigns more probability to siblings with
higher credit under the same suffix.

The empirical mean, standard error, and abstention statistic are
\begin{align}
\bar\Delta&=\frac1K\sum_k\Delta_k,\\
\operatorname{SE}(\Delta)&=
\sqrt{\frac{\sum_k(\Delta_k-\bar\Delta)^2}{K(K-1)}},\\
\operatorname{LCB}(s)&=\bar\Delta-z\operatorname{SE}(\Delta),
\qquad z=1.
\label{eq:supp-lcb}
\end{align}
The method accepts the local teacher when credits vary, agreement reaches the
threshold, and the LCB is positive; otherwise the probe produces no teacher.

For an accepted candidate, the confidence weight is
\begin{equation}
w(s)=\min\!\left(1,
\frac{\gamma(s)\operatorname{LCB}(s)}
{\max\{|\bar\Delta|+\operatorname{SE}(\Delta),10^{-12}\}}
\right).
\label{eq:supp-weight}
\end{equation}
Equation~\eqref{eq:supp-teacher} is solved again at the shrunken KL radius
$w(s)\delta$, yielding the stored teacher $\pi^*$.  The paired improvements and
LCB are recomputed for this final target.  Therefore weak but admissible
evidence produces a smaller redistribution rather than the same target with a
smaller loss coefficient.

\section{Component III: Bounded Credit Consolidation}

An accepted row stores $(s,\mathcal A_s,\pi^*,n_{\mathrm{score}}^{\mathrm{create}})$.
The queue contains at most 512 rows, and a row is active only while
\begin{equation}
n_{\mathrm{score}}-n_{\mathrm{score}}^{\mathrm{create}}<1000.
\label{eq:supp-ttl}
\end{equation}
Age is measured in physical terminal-score calls, not optimizer steps.  This
keeps the lifetime comparable when different arms spend different fractions of
their fixed budget on ordinary trajectories and probes.  Replaying a row does
not call the factor evaluator and does not recompute $\pi^*$ under a later
policy.

For active rows $\mathcal B$, define the student's sibling-restricted
distribution
\begin{equation}
\pf^\theta(a\mid s,\mathcal A_s)=
\frac{\pf^\theta(a\mid s)}
{\sum_{a'\in\mathcal A_s}\pf^\theta(a'\mid s)}.
\end{equation}
The auxiliary objective is
\begin{equation}
\begin{aligned}
\opd(\theta)={}&\frac1{|\mathcal B|}
\sum_{(s,\mathcal A_s,\pi^*)\in\mathcal B}\\
&\KL\!\left(\pi^*(\cdot\mid s)\|
\pf^\theta(\cdot\mid s,\mathcal A_s)\right).
\end{aligned}
\label{eq:supp-opd}
\end{equation}
For one row, ignoring the restriction Jacobian notation, the derivative with
respect to a sibling logit $z_i$ has the familiar cross-entropy form
\begin{equation}
\frac{\partial}{\partial z_i}
\KL(\pi^*\|\pf^\theta)=(\pf^\theta)_i-\pi_i^*.
\label{eq:supp-opd-gradient}
\end{equation}
The update therefore increases an underweighted accepted action and decreases
an overweighted one within the compared sibling set.

Let
$g_{\mathrm{ETB}}=\|\nabla_\theta\etb\|_2$ and
$g_{\mathrm{OPD}}=\|\nabla_\theta\opd\|_2$, measured on the forward-policy
parameters.  The applied scale is
\begin{equation}
\begin{aligned}
\lambda&=\min\!\left(10^4,
0.10\frac{g_{\mathrm{ETB}}}{g_{\mathrm{OPD}}+10^{-12}}\right),\\
\mathcal L&=\etb+\lambda\opd.
\end{aligned}
\label{eq:supp-balance}
\end{equation}
When the cap is inactive and $g_{\mathrm{OPD}}>0$, the auxiliary gradient norm
is approximately $0.10g_{\mathrm{ETB}}$.  When no accepted row is active,
$\opd$ contributes exactly zero.  This control bounds the local teacher's
optimization strength without replacing the global TB objective.

\section{Physical-Score Accounting and Matched Ablation}

One probe evaluates $m$ siblings under $K$ shared suffixes, so its direct cost
is
\begin{equation}
N_{\mathrm{score}}^{\mathrm{probe}}=mK=3\times4=12.
\label{eq:supp-probe-cost}
\end{equation}
Every terminal evaluation is charged, including rows later rejected by the
gate.  The formal budget is
\begin{equation}
N_{\mathrm{ordinary}}+N_{\mathrm{probe}}
=8008+1992=10000.
\label{eq:supp-total-budget}
\end{equation}
Since $1992/12=166$, the probe allocation permits exactly 166 complete probes.
Replay does not appear in Eq.~\eqref{eq:supp-total-budget} because it reuses
stored state-action targets and makes no evaluator call.

\begin{table}[t]
\centering
\small
\caption{Equal-physical-score ablation contract.}
\label{tab:supp-ablation}
\resizebox{\columnwidth}{!}{%
\begin{tabular}{lcccc}
\toprule
Method & Interface & Paired teacher & Gate & Consolidation \\
\midrule
Base Model & -- & -- & -- & -- \\
Components I--II (Gate Off) & $\checkmark$ & $\checkmark$ & -- & -- \\
Components I--II & $\checkmark$ & $\checkmark$ & $\checkmark$ & -- \\
Components I--III & $\checkmark$ & $\checkmark$ & $\checkmark$ & $\checkmark$ \\
\bottomrule
\end{tabular}
}
\end{table}

All OPD arms use the same $8008/1992$ allocation.  Gate Off retains the same
visited-state interface, siblings, four matched suffixes, teacher temperature,
and KL radius, while disabling winner agreement, positive-LCB abstention, and
confidence shrinkage.  The Gate-Off to Gate-On contrast therefore evaluates
the reliability rule beyond paired evaluation alone.  Components I--II use an
accepted row once at the next optimizer update.  Components I--III retain the
same acceptance rule and change only bounded persistence and measured gradient
strength, isolating the consolidation mechanism.

\section{Implementation and Reproduction Contract}

\begin{table}[t]
\centering
\small
\caption{Final training and OPD configuration.}
\label{tab:supp-hparams}
\resizebox{\columnwidth}{!}{%
\begin{tabular}{ll}
\toprule
Item & Value \\
\midrule
Optimizer / learning rate & Adam / $10^{-4}$ \\
Encoder & 2-layer RGCN, hidden size 128 \\
Update frequency & 128 sampled trajectories \\
Maximum expression length & 20 \\
Pool capacity & 50 \\
Entropy coefficient / temperature & 0.01 / 1.0 \\
Total / ordinary / probe scores & 10000 / 8008 / 1992 \\
Sibling actions / shared suffixes & 3 / 4 \\
Teacher temperature / target KL & 1.0 / 0.03 \\
Agreement / LCB multiplier & 0.75 / 1.0 \\
Replay capacity / score TTL & 512 / 1000 \\
Target OPD gradient ratio & 0.10 \\
Random seeds & 0, 1, 2 \\
Hardware & NVIDIA A100 40GB; PyTorch \\
\bottomrule
\end{tabular}
}
\end{table}

The expression search uses OPEN, CLOSE, HIGH, LOW, and VOLUME, with windows
$\{10,20,30,40,50\}$.  Its constants are $-30$, $-10$, $-5$, $-2$, $-1$,
$-0.5$, $-0.01$, $0.01$, $0.5$, $1$, $2$, $5$, $10$, and $30$.  Unary,
binary, and time-series operators follow the accompanying anonymous code
archive.  Training uses the training-period targets specified below.

The Chinese universes use 2010--2020 for training, 2021 for validation, and
2022--2024 for test.  The S\&P 500 uses 2010--2016, 2017, and
2018--2020, respectively.  The target is the 20-trading-day forward close
return, \texttt{Ref(close,-20)/close-1}.  Validation-Sign Equal-50 fixes each
factor direction using validation Pearson IC and assigns magnitude $1/50$
before one untouched test evaluation.

\section{Evaluation Metrics}

For dates $t=1,\ldots,T$ and cross-sectional observations $j=1,\ldots,n_t$,
let $f_{tj}$ be the factor score and $y_{tj}$ the forward return.  The per-date
Pearson information coefficient and its reported time-series mean are
\begin{equation}
\mathrm{IC}_t=\operatorname{Corr}_{j}(f_{tj},y_{tj}),
\qquad
\mathrm{IC}=\frac1T\sum_{t=1}^{T}\mathrm{IC}_t.
\end{equation}
Writing $\bar f_t=n_t^{-1}\sum_j f_{tj}$ and
$\bar y_t=n_t^{-1}\sum_j y_{tj}$, the cross-sectional correlation is
\begin{equation}
\operatorname{Corr}_{j}(f_{tj},y_{tj})=
\frac{\sum_j(f_{tj}-\bar f_t)(y_{tj}-\bar y_t)}
{\sqrt{\sum_j(f_{tj}-\bar f_t)^2}
 \sqrt{\sum_j(y_{tj}-\bar y_t)^2}}.
\end{equation}
RankIC applies the same calculation after independently replacing factor scores
and returns by their within-date ranks, using average ranks for ties:
\begin{equation}
\begin{aligned}
\mathrm{RankIC}_t={}&
\operatorname{Corr}_{j}(\operatorname{rank}(f_{tj}),
\operatorname{rank}(y_{tj})),\\
\mathrm{RankIC}={}&\frac1T\sum_{t=1}^{T}\mathrm{RankIC}_t.
\end{aligned}
\end{equation}
With sample standard deviations $s_{\mathrm{IC}}$ and
$s_{\mathrm{RankIC}}$ across dates, the information ratios are
\begin{equation}
\mathrm{ICIR}=\frac{\mathrm{IC}}{s_{\mathrm{IC}}},
\qquad
\mathrm{RankICIR}=\frac{\mathrm{RankIC}}{s_{\mathrm{RankIC}}}.
\end{equation}

For factor $m\in\{1,\ldots,50\}$, validation Pearson IC fixes
$d_m=\operatorname{sign}(\mathrm{IC}^{\mathrm{valid}}_m)$.  The frozen
Validation-Sign Equal-50 score is
\begin{equation}
F_{tj}=\frac1{50}\sum_{m=1}^{50}d_m f_{mtj}.
\end{equation}
The portfolio mapping described in the main paper converts $F_t$ into a daily
return sequence $r_1,\ldots,r_T$.  With $A$ trading periods per year and wealth
$W_t=\prod_{u=1}^{t}(1+r_u)$, the portfolio metrics are
\begin{align}
\mathrm{AR}&=W_T^{A/T}-1,\\
\mathrm{SR}&=\sqrt{A}\,\frac{\bar r}{s_r},\\
\mathrm{MDD}&=\max_{1\le u\le v\le T}
\left(1-\frac{W_v}{W_u}\right),
\end{align}
where $\bar r=T^{-1}\sum_t r_t$ and $s_r$ is the sample standard deviation of
daily returns.  MDD is reported as a positive drawdown magnitude.

For $S$ random seeds and any reported statistic $z_s$, tables display
\begin{equation}
\bar z=\frac1S\sum_{s=1}^{S}z_s,
\qquad
s_z=\sqrt{\frac1{S-1}\sum_{s=1}^{S}(z_s-\bar z)^2},
\end{equation}
with $s_z$ shown in parentheses below the mean.


\end{document}